\documentclass[journal]{IEEEtran}
\usepackage{graphicx}
\usepackage{amsmath, amssymb}
\usepackage{caption, subcaption}
\usepackage{float}
\usepackage{multirow}
\usepackage{hyperref}
\usepackage{soul,color}
\begin{document}
\title{3D/2D regularized CNN feature hierarchy for Hyperspectral image classification}
\author{Muhammad Ahmad, Manuel Mazzara, and Salvatore Distefano
\thanks{M. Ahmad is with the Department of Computer Science, National University of Computer and Emerging Sciences, Islamabad, Chiniot-Faisalabad Campus, Chiniot 35400, Pakistan}
\thanks{M. Mazzara is with Innopolis University, Innopolis 420500, Russia}
\thanks{S. Distefano is with  Dipartimento di Matematica e Informatica---MIFT, University of Messina, Messina 98121, Italy}
}
\markboth{Preprint Submitted to arXiv, April~2021}
{M.Ahmad \MakeLowercase{\textit{et al.}}: 3D/2D regularized CNN}
\maketitle
\begin{abstract}
Convolutional Neural Networks (CNN) have been rigorously studied for Hyperspectral Image Classification (HSIC) and are known to be effective in exploiting joint spatial-spectral information with the expense of lower generalization performance and learning speed due to the hard labels and non-uniform distribution over labels. Several regularization techniques 
have been used to overcome the aforesaid issues. 
However, sometimes models learn to predict the samples extremely confidently which is not good from a generalization point of view. Therefore, this paper proposed an idea to enhance the generalization performance of a hybrid CNN for HSIC using soft labels that are a weighted average of the hard labels and uniform distribution over ground labels. The proposed method helps to prevent CNN from becoming over-confident. We empirically show that in improving generalization performance, label smoothing also improves model calibration which significantly improves beam-search. Several publicly available Hyperspectral datasets are used to validate the experimental evaluation which reveals improved generalization performance, statistical significance, and computational complexity as compared to the state-of-the-art models. 
The source code will be made available at https://github.com/mahmad00.
\end{abstract}
\begin{IEEEkeywords}
Beam-search; Label Smoothing; Hybrid Convolutional Neural Network (CNN); Hyperspectral Images Classification (HSIC).
\end{IEEEkeywords}
\IEEEpeerreviewmaketitle
\section{Introduction}
\label{Sec.1}

\IEEEPARstart{H}{yperspectral} Imaging (HSI) has been extensively utilized for many real-world applications \cite{alcolea20}; for instance, Crop monitoring, vegetation coverage, precision agriculture, land resources, oil spills, water quality, meat adulteration \cite{ayaz2020myoglobin, ayaz2020myoglobin1}, adulteration in house hold products such as color adulteration detection in red chili \cite{Khan2020}, microbial spoilage and shelf-life of bakery products \cite{Zainab2020}. 

Thus, HSI Classification (HSIC) has received remarkable attention and intensive research results have been reported in the past few decades \cite{ahmad2021hyperspectral}. According to the literature, HSIC can be categorized into spatial, spectral, and spatial-spectral feature methods \cite{9330782}. The spectral feature can be labeled as a primitive characteristic of HSI also known as spectral curve or vector whereas the spatial feature contains the relationship between the central pixel and its context which significantly uplifts the performance \cite{shabbir2021hyperspectral}. 

In the last few years, deep learning especially Convolutional Neural Network (CNN) has received widespread attention due to its ability to automatically learn non-linear features for classification, i.e., overcome the challenges of hand-crafted features for classification. Moreover, it can jointly investigate the spatial-spectral information and such models can be categorized into two groups, i.e., single and two-stream, more information regarding single or two-stream methods can be found in \cite{JIA2021179}. This work explicitly investigate a  single-stream method but in a combination of 3-D and 2-D layers. 

Irrespective of the single or two-stream method, all deep learning frameworks are sensitive to the loss which needs to be minimized \cite{104451, sam11489}. Several classical works showed that the gradient descent to minimize cross-entropy performs better in terms of classification and has fast convergence, however, to some extent, leads to the overfitting. Several regularization techniques such as dropout, L1, L2, etc., have been used to overcome the overfitting issues together with several other exotic objectives performed exceptionally well than the standard cross-entropy \cite{9324175}. Recently, a work \cite{inproceedings} proposed a Label Smoothing (LS) technique that improves the accuracy significantly by computing cross-entropy with a weighted mixture of targets with uniform distribution instead of hard-coded targets. 

Since then, LS has been known to improve the classification performance of deep models \cite{1890854}. However, the original idea was used to improve the classification performance of only the inception model on ImageNet data \cite{inproceedings}. However, since then, various image classification models (training models) have used LS \cite{Le_2019, ZophVSL17}. Though the LS technique is widely used trick to improve the classification performance and to speed up the convergence process, however, it has not been much explored for HSIC, and above all, it has not been much explored regarding when and why LS should work. This is due to, in HSIC, sometimes models learn to predict the training samples extremely confidently which is not good from a generalization point of view. 

Therefore, this paper proposed an idea to enhance the generalization performance of a hybrid CNN for HSIC using soft labels that are a weighted average of the hard labels and uniform distribution over target labels. The proposed method helps to prevent Hybrid CNN from becoming over-confident. We empirically show that in improving generalization performance, LS also improves model calibration which significantly improves beam-search. Several publicly available Hyperspectral datasets are used to validate the experimental evaluation which reveals improved generalization performance, statistical significance, and computational complexity as compared to the state-of-the-art 2-D/3-D CNN models.


\section{Proposed Methodology}
\label{Sec.2}

Let us assume that the Hyperspectral data can be represented as $R^{(M \times N) \times B^{*}} = [r_1, r_2, r_3, \dots, r_S]^T$, where $B^{*}$ be the total number of bands and $(M \times N)$ are the samples per band belonging to $Y$ classes and $r_i = [r_{1,i},~r_{2,i},~r_{3,i},~\dots, r_{B^{*},i}]^T$ is the $i^{th}$ sample in the Hyperspectral Data. Suppose $(r_i, y_i) \in (\mathcal{R}^{ M \times N \times B^{*}} , \mathcal{R}^Y)$, where $y_i$ is the class label of the $i^{th}$ sample.
Later instead of using the "hot encoding" in hybrid model, we introduce a smoothing technique $\mu(y|r_i)$. Thus the new ground truths $(r_i, y_i)$ would be:

\begin{equation}
    p^{\prime}(y|r_i) = (1-\varepsilon) p(y|r_i) + \varepsilon \mu(y|r_i)
\end{equation}

\begin{equation}
    f(x)= 
\begin{cases}
    1 - \varepsilon + \varepsilon \mu(y|r_i) & if ~ y = y_i \\
    \varepsilon \mu(y|x_i)              & otherwise
\end{cases}
\end{equation}
where $\varepsilon \in [0, 1]$ is a weight factor, and note that $\sum_{y=1}^{Y} p^{\prime}(y|r_i) = 1$. These new ground truths has been used in loss function instead of hot-encoding. 

\begin{equation}
L^{\prime} = - \sum_{i=1}^{M \times N} \sum_{y=1}^{Y} p^{\prime}(y|r_i) \log q_{\theta}(y|r_i)
\end{equation}

\begin{equation}
    L^{\prime} = - \sum_{i=1}^{M \times N} \sum_{y=1}^{Y} \big[ (1-\varepsilon) p(y|r_i) + \varepsilon \mu(y|r_i) \big] \log q_{\theta}(y|r_i)
\end{equation}

\begin{equation}
\begin{aligned}
L^{\prime} = \sum_{i=1}^{M \times N} \bigg\{(1-\varepsilon) \Big[ - \sum_{y=1}^{Y} p(y|r_i) \log q_{\theta}(y|r_i) \Big] + \\
\varepsilon \Big[ -  \sum_{y=1}^{Y} \mu(y|x_i) \log q_{\theta}(y|r_i) \Big] \bigg\}
\end{aligned}
\end{equation}

\begin{equation}
    L^{\prime} = \sum_{i=1}^{M \times N} \Big[ (1-\varepsilon) H_i(p, q_{\theta}) + \varepsilon H_i(u, q_{\theta}) \Big]
    \label{ABC}
\end{equation}

One can observe that each ground truth, the loss contribution is a mixture of entropy between predicted distribution ($H_i(p, q^{\theta})$) and the hot-encoding, and the entropy between the predicted distribution ($H_i(\mu, q^{\theta})$) and the noise distribution. While training, $H_i(p, q^{\theta}) = 0$ if the model learns to predict the distribution confidently, however, $H_i(\mu, q^{\theta})$ will increase dramatically. To overcome this phenomenon, we used a regularizer $H_i(\mu, q^{\theta})$ to prevent the model from predicting too confidently. In practice, $\mu(y|r)$ is a uniform distribution that does not dependant on Hyperspectral Data. That is to say $\mu(y|r) = \frac{1}{Y}$. The ultimate network is trained for $50$ epochs using a mini-batch size of 256 without any batch normalization and data augmentation. In a nutshell, the details of 3D/2D convolutional layers and kernels are as follows: $3D\_conv\_layer\_1 = 8 \times 5 \times 5 \times 7 \times 1$ i.e. $K_{1}^{1} = 5, K_{2}^{1} = 5$ and $K_{3}^{1} = 7$. $3D\_conv\_layer\_2 = 16 \times 5 \times 5 \times 5 \times 8$  i.e. $K_{1}^{2} = 5, K_{2}^{2} = 5$. $K_{3}^{2} = 5$. $3D\_conv\_layer\_3 = 32 \times 3 \times 3 \times 3 \times 16$ i.e. $K_{1}^{3} = 3, K_{2}^{3}= 3$ and $K_{3}^{3} = 3$. $3D\_conv\_layer\_4 = 64 \times 3 \times 3 \times 3 \times 32$ i.e. $K_{1}^{3} = 3, K_{2}^{3}= 3$ and $K_{3}^{3} = 3$. $2D\_conv\_layer\_5 = 128 \times 3 \times 3 \times 64$ i.e. $K_{1}^{2} = 3$ and $K_{2}^{2}= 3$. Three 3D convolutional layers are employed to increase the number of spectral-spatial feature maps and one 2D convolutional layer is used to discriminate the spatial features within different spectral bands while preserving the spectral information. Initially, the weights are randomized and then optimized using back-propagation with the Adam optimizer by using the loss function presented in equation \ref{ABC}. Further details regarding the Hybrid CNN architecture in terms of types of layers, dimensions of output feature maps and number of trainable parameters can be found in \cite{ahmad2021hyperspectral}.

\section{Experimental Settings and Results}
\label{Sec.3}
The experiments have been conducted on two real HSI datasets, namely, Indian Pines (IP), and Pavia University (PU). These datasets are acquired by two different sensors i.e, Reflective Optics System Imaging Spectrometer (ROSIS) and Airborne Visible/Infrared Imaging Spectrometer (AVIRIS) \cite{ahmad2021hyperspectral}. The experimental results explained in this work have been obtained through Google Colab \cite{carneiro2018} which is an online platform to execute any python environment while having a good internet speed to execute the code. Google Colab provides the option to execute many versions of python, Graphical Process Unit (GPU), upto 358$+$ GB of cloud storage, and upto 25 GB of Random Access Memory (RAM). 

\begin{figure}[!hbt]
	\begin{subfigure}{0.24\textwidth}
		\includegraphics[width=0.99\textwidth]{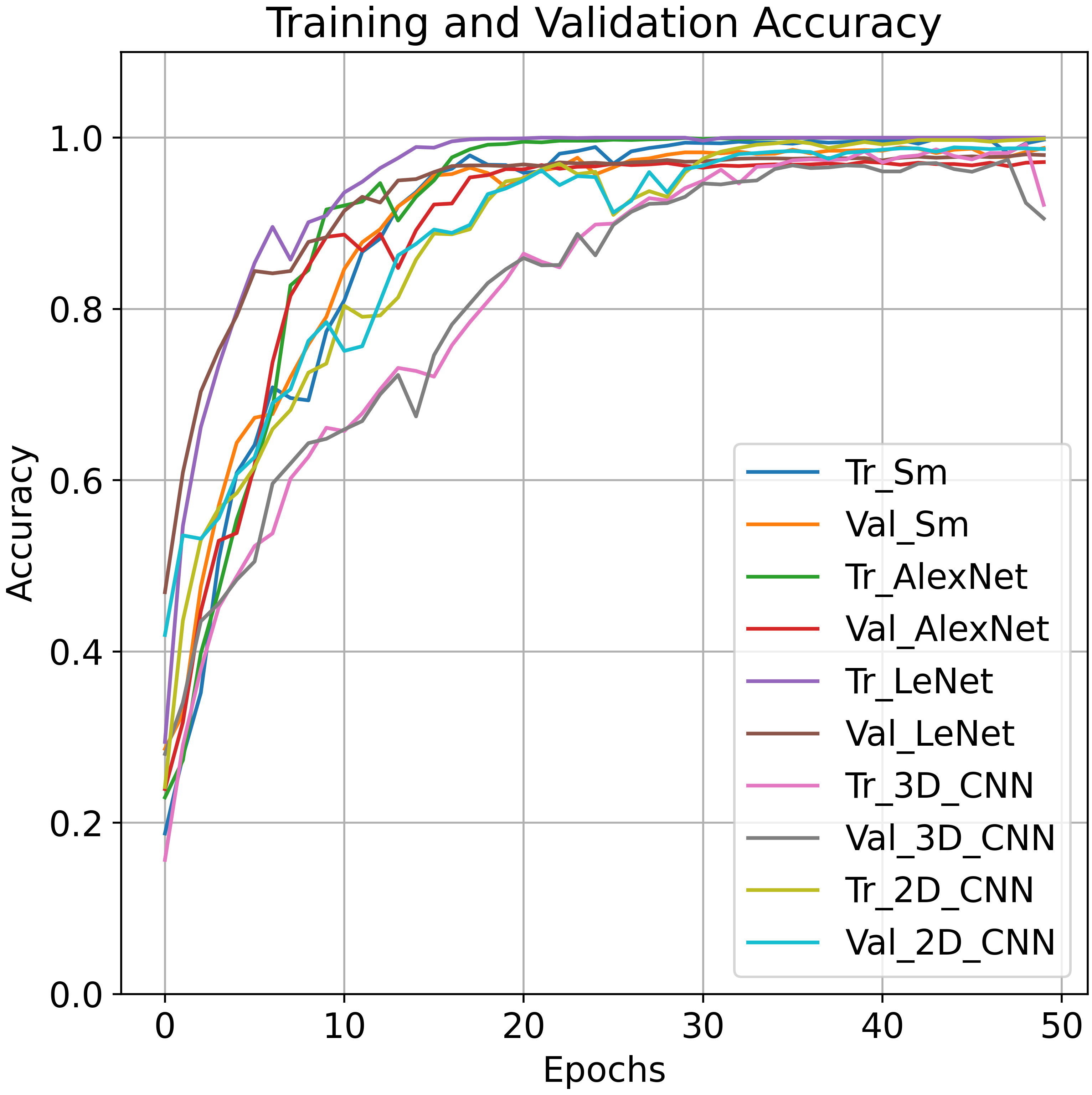}
		\centering
		\caption{Accuracy} 
		\label{Fig1A}
	\end{subfigure}
	\begin{subfigure}{0.24\textwidth}
		\includegraphics[width=0.99\textwidth]{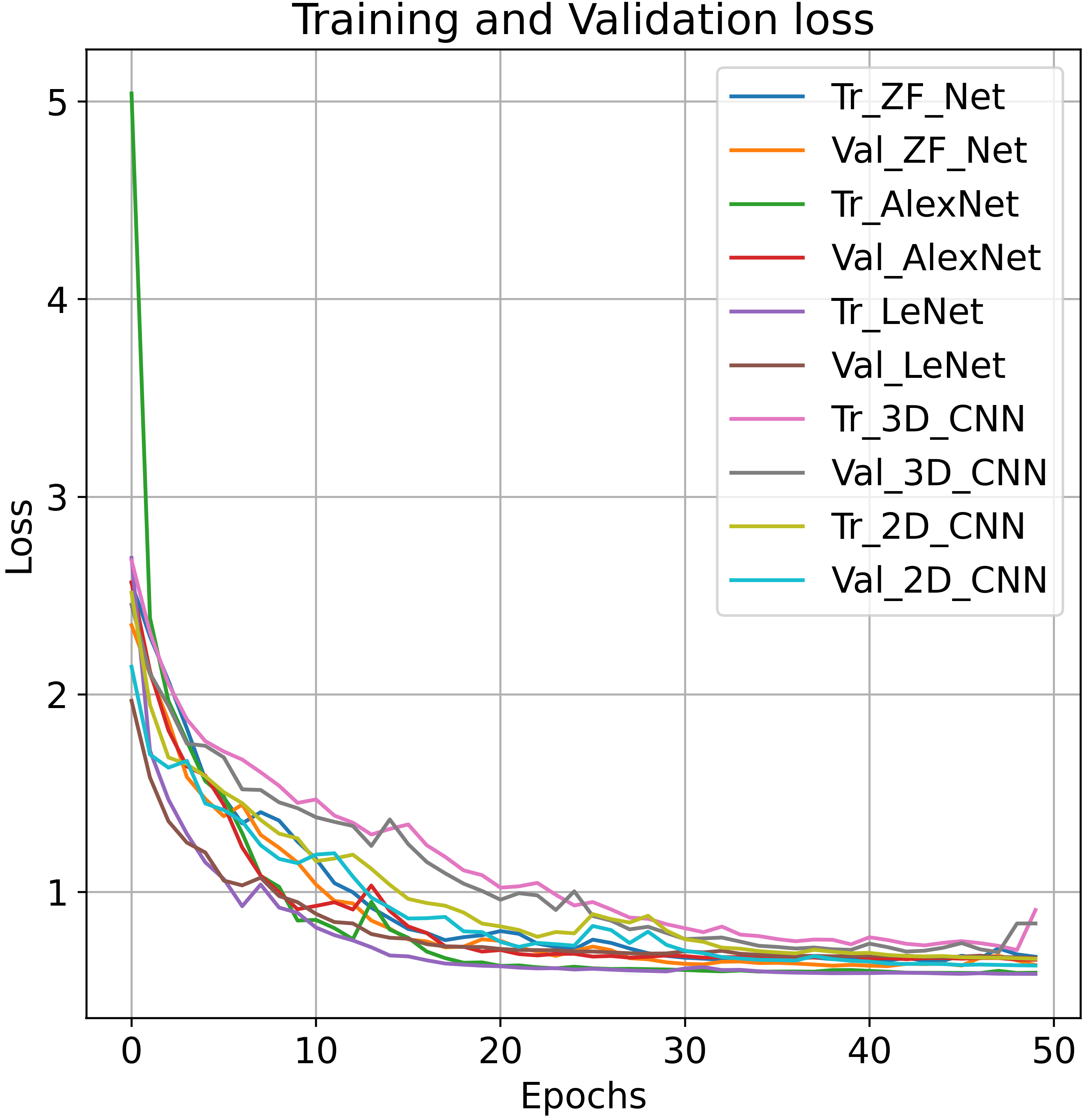}
		\centering
		\caption{Loss}
		\label{Fig1B}
	\end{subfigure}
\caption{Accuracy and Loss for Training  and  Validation  sets on Indian Pines for $50$ epochs.}
\label{Fig.1}
\end{figure}

In all the experiments, the initial size of the train/validation/test sets is set to 25\%/25\%/50\% to validate the proposed model as well as several other state-of-art-deep models. The comparative methods include AlexNet 
\cite{9362309}, LeNet 
\cite{9349481}, 2D CNN 
\cite{9381397}, and 3D CNN 
\cite{9307220} models. Moreover, the following accuracy metrics 
have been conducted to validate the claims made in this manuscript. The accuracy metrics include Kappa ($\kappa$)\footnote{$\kappa$ is known as a statistical metric that considered the mutual information regarding a strong agreement among classification and ground-truth maps}, Average (AA)\footnote{AA represents the average class-wise classification performance}, and Overall (OA)\footnote{OA is computed as the number of correctly classified examples out of the total test examples}.
All these metrics are computed using the following equations. 

\begin{equation}
    Kappa ~(\kappa) = \frac{P_o - P_e}{1 - P_e}
\end{equation}
where 
\begin{equation*}
    P_e = P^+ + P^-
\end{equation*}

\begin{equation*}
P^+ = \frac{TP + FN}{TP + FN + FP + TN} \times \frac{TP + FN}{TP + FN + FP + TN}
\end{equation*}

\begin{equation*}
P^- = \frac{FN + TN}{TP + FN + FP + TN} \times \frac{FP + TN}{TP + FN + FP + TN}
\end{equation*}

\begin{equation*}
    P_o = \frac{TP + TN}{TP + FN + FP + TN}
\end{equation*}

\begin{equation}
    Overall ~(OA) = \frac{1}{K} \sum_{i = 1}^K TP_i
\end{equation}
where $TP$ and $FP$ are true and false positive, $TN$ and $FN$ are true and false negative, respectively. For fair comparison purposes, the learning rate for all these models including hybrid models is set to 0.001, Relu as the activation function for all layers except the output layer on which Softmax is used, patch size is set of 15, and for all the experiments, 15 most informative bands have been selected using principal component analysis to reduce the computational load. The convergence accuracy and loss of our proposed pipeline and all other competing methods for 50 epochs is presented in Figure \ref{Fig.1}. From loss and accuracy curves, one can conclude that the smoothing has faster convergence than the model without convergence.

\subsection{Indian Pines}

\begin{figure}[!hbt]
    \centering
	\begin{subfigure}{0.070\textwidth}
		\includegraphics[width=0.99\textwidth]{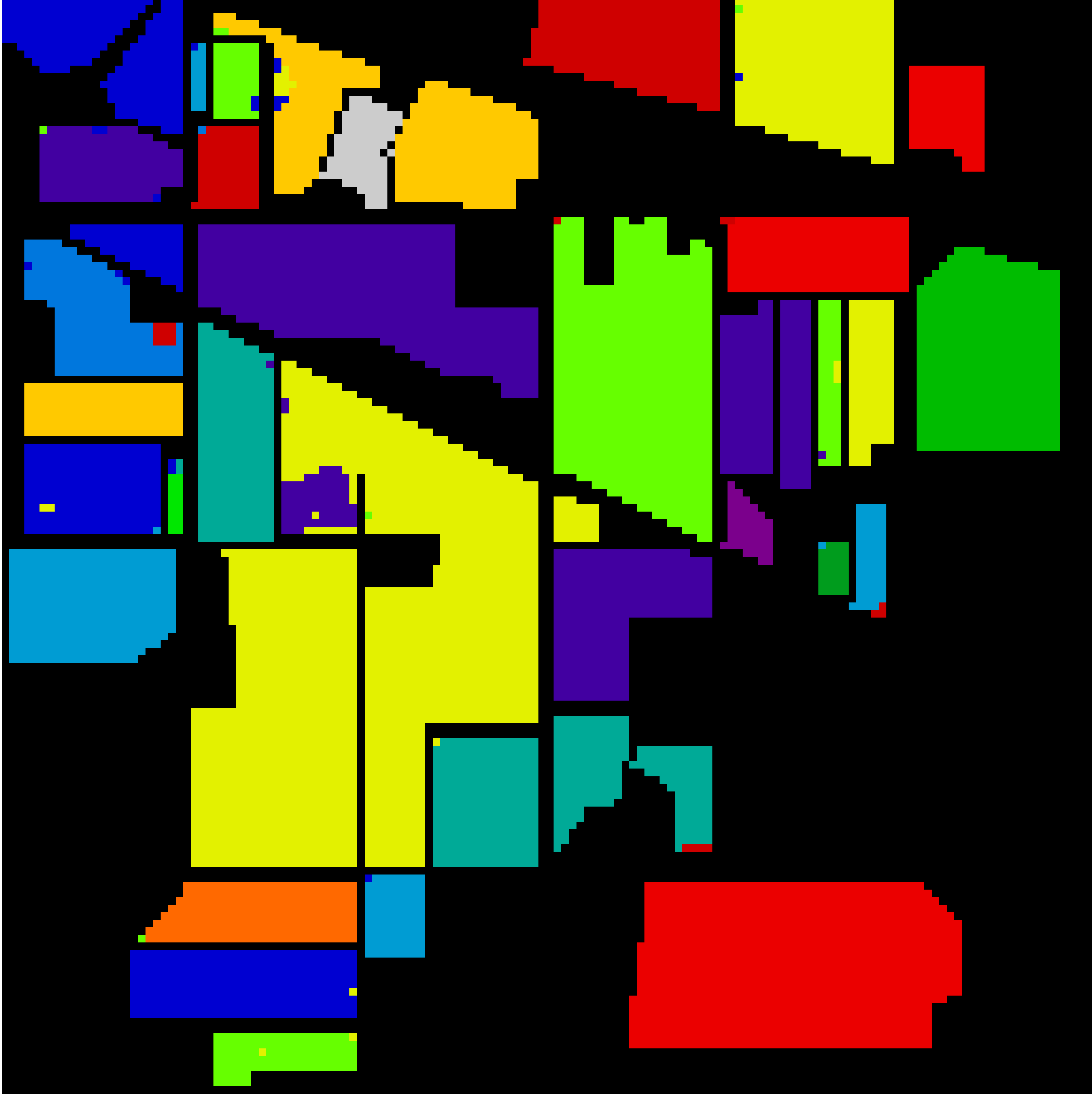}
		\caption{2D} 
		\label{Fig2A}
	\end{subfigure}
	\begin{subfigure}{0.070\textwidth}
		\includegraphics[width=0.99\textwidth]{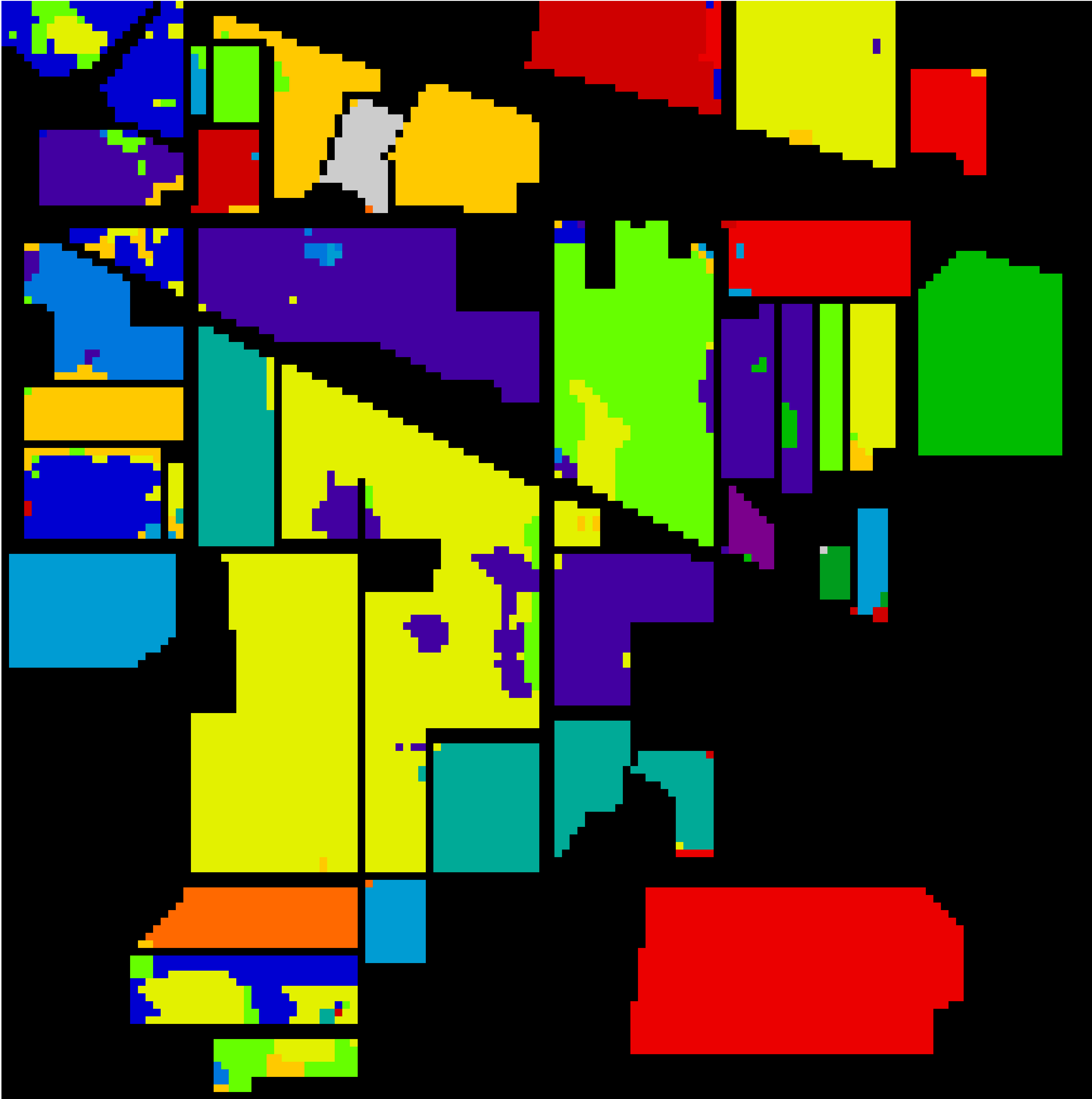}
		\caption{3D}
		\label{Fig2B}
	\end{subfigure}
		\begin{subfigure}{0.070\textwidth}
		\includegraphics[width=0.99\textwidth]{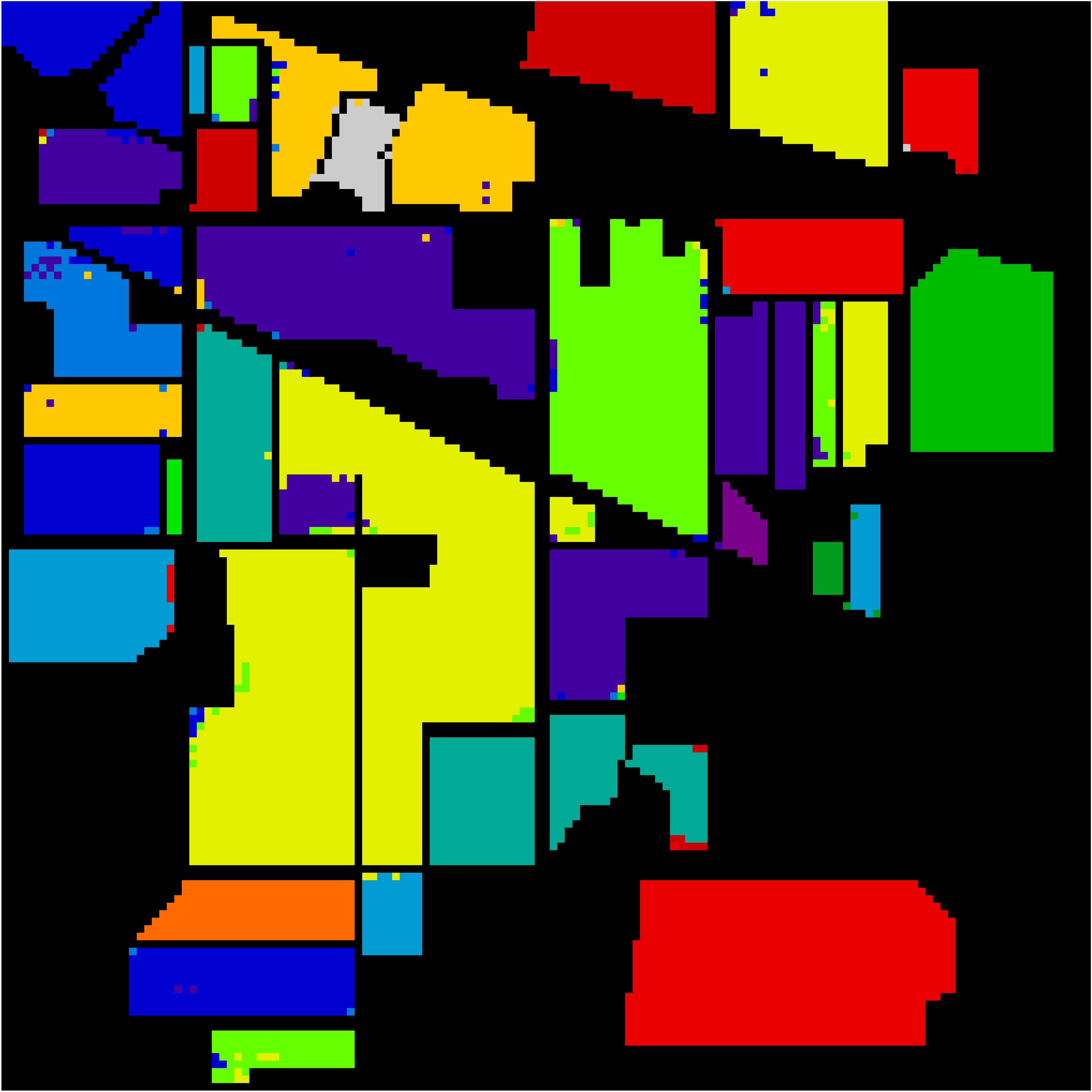}
		\caption{A.Net} 
		\label{Fig2C}
	\end{subfigure}
	\begin{subfigure}{0.070\textwidth}
		\includegraphics[width=0.99\textwidth]{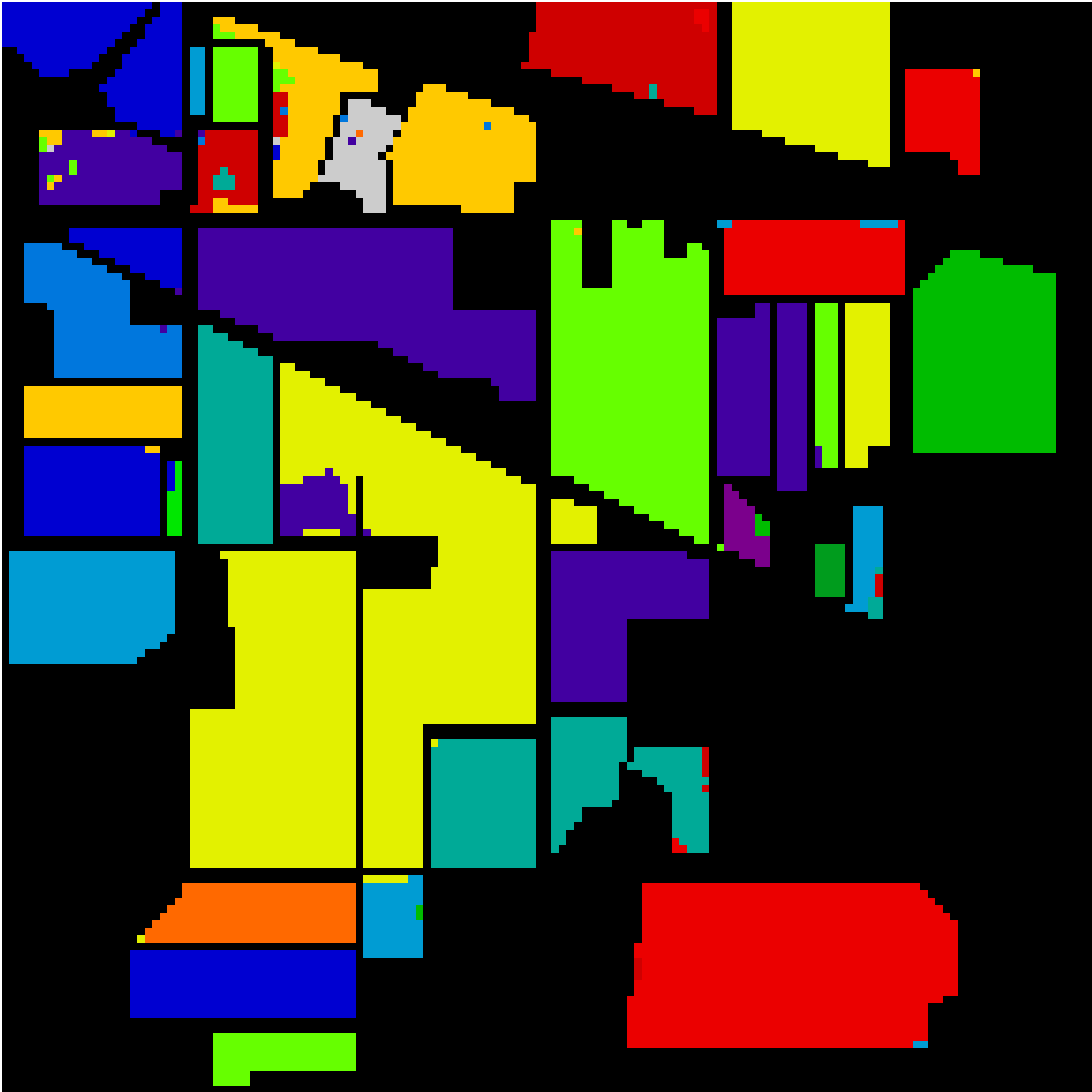}
		\caption{LeNet}
		\label{Fig2D}
	\end{subfigure}
		\begin{subfigure}{0.070\textwidth}
		\includegraphics[width=0.99\textwidth]{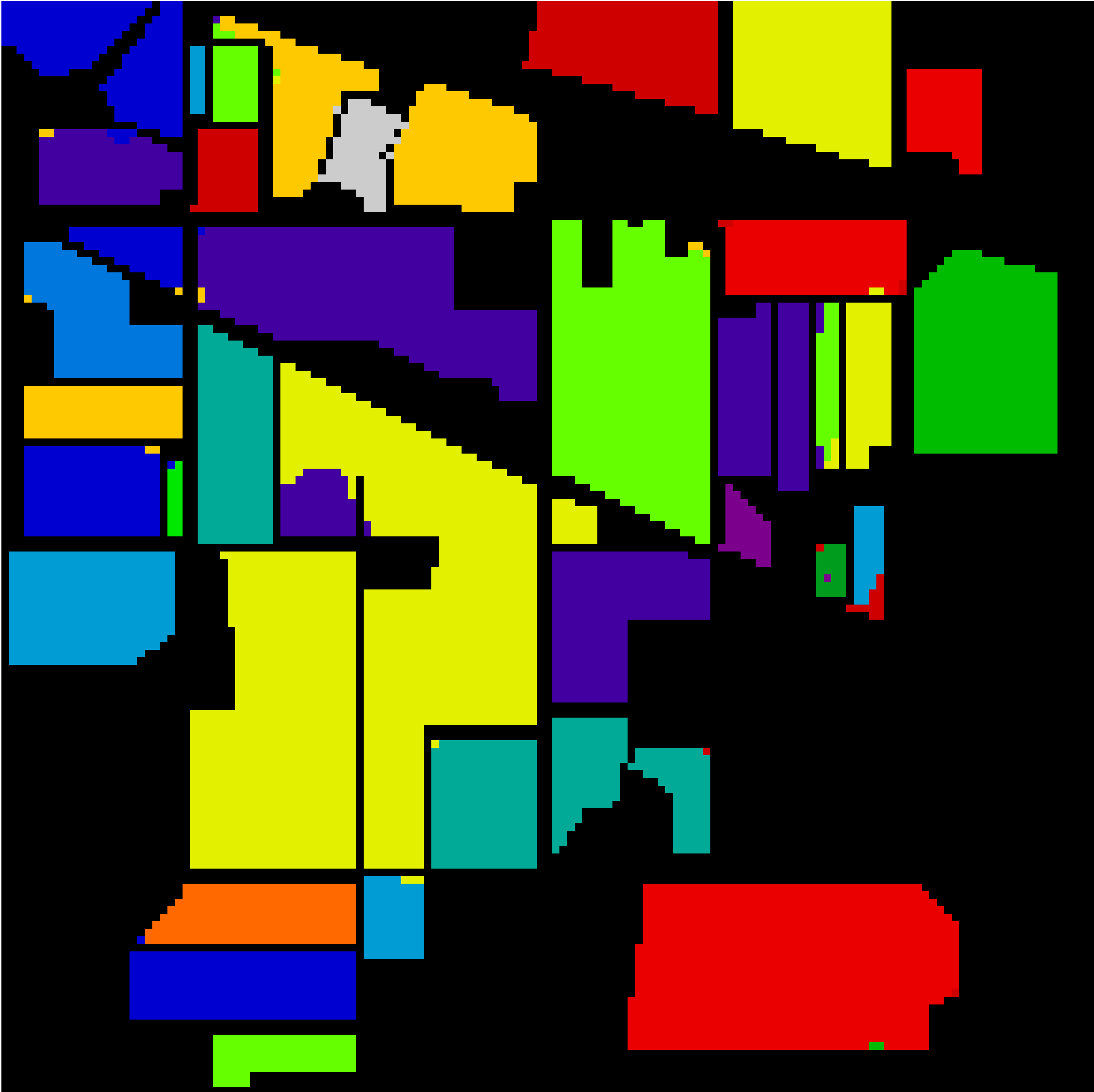}
		\caption{WoS} 
		\label{Fig2E}
	\end{subfigure}
	\begin{subfigure}{0.070\textwidth}
		\includegraphics[width=0.99\textwidth]{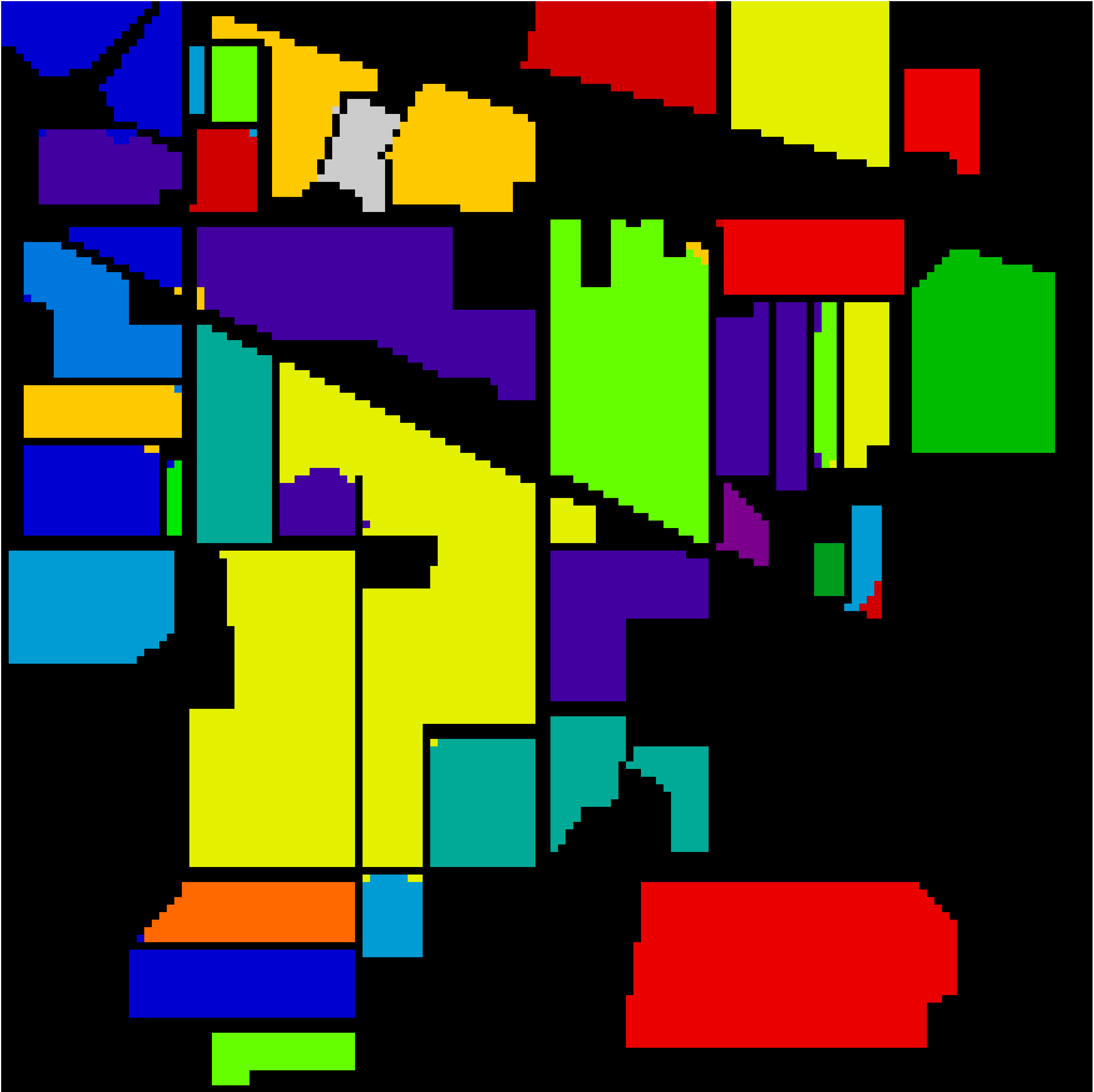}
		\caption{WS}
		\label{Fig2F}
	\end{subfigure}
\caption{\textbf{Indian Pines:} Classification accuracy: Fig. \ref{Fig2A}: 2D-CNN = 98.94\%, Fig. \ref{Fig2B}: 3D CNN = 91.57\%, Fig. \ref{Fig2C}: AlexNet = 97.65\%, Fig. \ref{Fig2D}: LeNet = 98.14\%, Fig. \ref{Fig2E}: WoS (Hybrid net without Smoothing) = 98.98\%, and Fig. \ref{Fig2F} WS (with Smoothing) = \textbf{99.29\%}.}
\label{Fig.2}
\end{figure}

Indian pines (IP) dataset is acquired using AVIRIS sensor over the northwestern Indiana test site. IP data consist of $145 \times 145$ spatial dimensions and 224 spectral dimensions with total of 16 classes in which all are not mutually exclusive. Some of the water absorption bands are removed and the remaining 200 bands are used for the experimental process. This data is consists of 2/3 agriculture, 1/3 forest, and other vegetation. Less than 5\% of total coverage consists of crops that are in an early stage of growth. Building, low-density housing, two dual-lane highway, small roads, and a railway line are also a part of it. Further details about the experimental datasets can be found at \cite{Grupointel}. Table \ref{Tab.1} and Figure \ref{Fig.2} presents in-depth comparative accuracy analysis on the IP dataset.

\begin{table}[!hbt]
    \centering
    \caption{\textbf{Indian Pines:} Performance analysis of different state-of-the-art models. The higher accuracies are emphasised.}
    \resizebox{\columnwidth}{!}{
    \begin{tabular}{cccccccc} \\ \hline 
        \textbf{Class} & \textbf{Train/Val/Test} & \textbf{2D} & \textbf{3D} & \textbf{AlexNet} & \textbf{LeNet} & \textbf{WoS} & \textbf{WS} \\ \hline 
        Alfalfa & 11/12/23 & 100 & 91.3043 & 95.6521 & 82.6086 & 100 & \textbf{100} \\ 
        Corn-notill & 357/357/714 & 98.3193 & 93.5574 & 97.3389 & 97.0588 & 98.4593 & \textbf{98.8795} \\  
        Corn-mintill & 207/208/415 & 99.5180 & 66.7469 & 98.3132 & 99.5180 & 99.5180 & \textbf{99.5180} \\ 
        Corn & 59/59/118 & 94.0677 & 90.6779 & 93.2203 & 99.1525 & 100 & \textbf{100} \\ 
        Grass-pasture & 121/121/242 & 98.3471 & 97.1074 & 96.2809 & 94.2148 & 95.0413 & \textbf{96.2809} \\ 
        Grass-trees & 182/183/365 & 98.9041 & 97.5342 & 98.3561 & 98.9041 & 99.4520 & \textbf{99.7260} \\ 
        Grass-mowed & 7/7/14 & 92.8571 & 92.8571 & 100 & 100 & 92.8571 & \textbf{100} \\ 
        Hay-windrowed & 119/120/239 & 100 & 100 & 100 & 100 & 100 & \textbf{100} \\ 
        Oats & 5/5/10 & 70 & 0 & 100 & 70 & 100 & \textbf{100} \\
        Soybean-notill & 243/243/486 & 98.5596 & 82.3045 & 93.6213 & 99.3827 & 97.9423 & \textbf{97.9423} \\
        Soybean-mintill & 614/614/1228 & 99.6742 & 92.4267 & 97.8013 & 99.9185 & 99.7557 & \textbf{99.8371} \\
        Soybean-clean & 148/149/297 & 97.6430 & 98.6531 & 96.9696 & 95.2861 & 98.3164 & \textbf{99.6632} \\
        Wheat & 51/51/102 & 99.0196 & 98.0392 & \textbf{100} & 99.0196 & 99.0196 & 99.0196 \\ 
        Woods & 316/317/633 & 99.8420 & 99.3680 & 99.5260 & 98.8941 & 99.2101 & \textbf{99.8420} \\ 
        Buildings & 96/97/193 & 99.4818 & 90.6735 & \textbf{100} & 91.1917 & 100 & 99.4818 \\ 
        Stone-Steel & 23/23/46 & 100 & 97.8260 & 100 & 93.4782 & 100 & \textbf{100} \\ \hline 
        \multicolumn{2}{c}{\textbf{Training Time}} & 55.6695 & 250.1662 & 919.5566 & 61.8763 & 250.1622 & 248.5993 \\ \hline 
        \multicolumn{2}{c}{\textbf{Test Time}} & 1.4897 & 4.0402 & 5.6891 & 1.2752 & 4.6499 & 3.9997 \\ \hline 
        \multicolumn{2}{c}{\textbf{Overall Accuracy}} & 98.9463 & 91.5707 & 97.6585 & 98.14634 & 98.9853 & \textbf{99.2975} \\ \hline
        \multicolumn{2}{c}{\textbf{Average Accuracy}} & 98.7980 & 86.8173 & 97.9425 & 94.9142 & 98.7232 & \textbf{99.3869}\\ \hline
        \multicolumn{2}{c}{\textbf{Kappa ($\kappa$)}} & 96.6396 & 90.3561 & 97.3312 & 97.8853 & 98.8430 & \textbf{99.1990} \\ \hline
    \end{tabular}}
    \label{Tab.1}
\end{table}

\subsection{Pavia University}

\begin{figure}[!hbt]
    \centering
	\begin{subfigure}{0.070\textwidth}
		\includegraphics[width=0.99\textwidth]{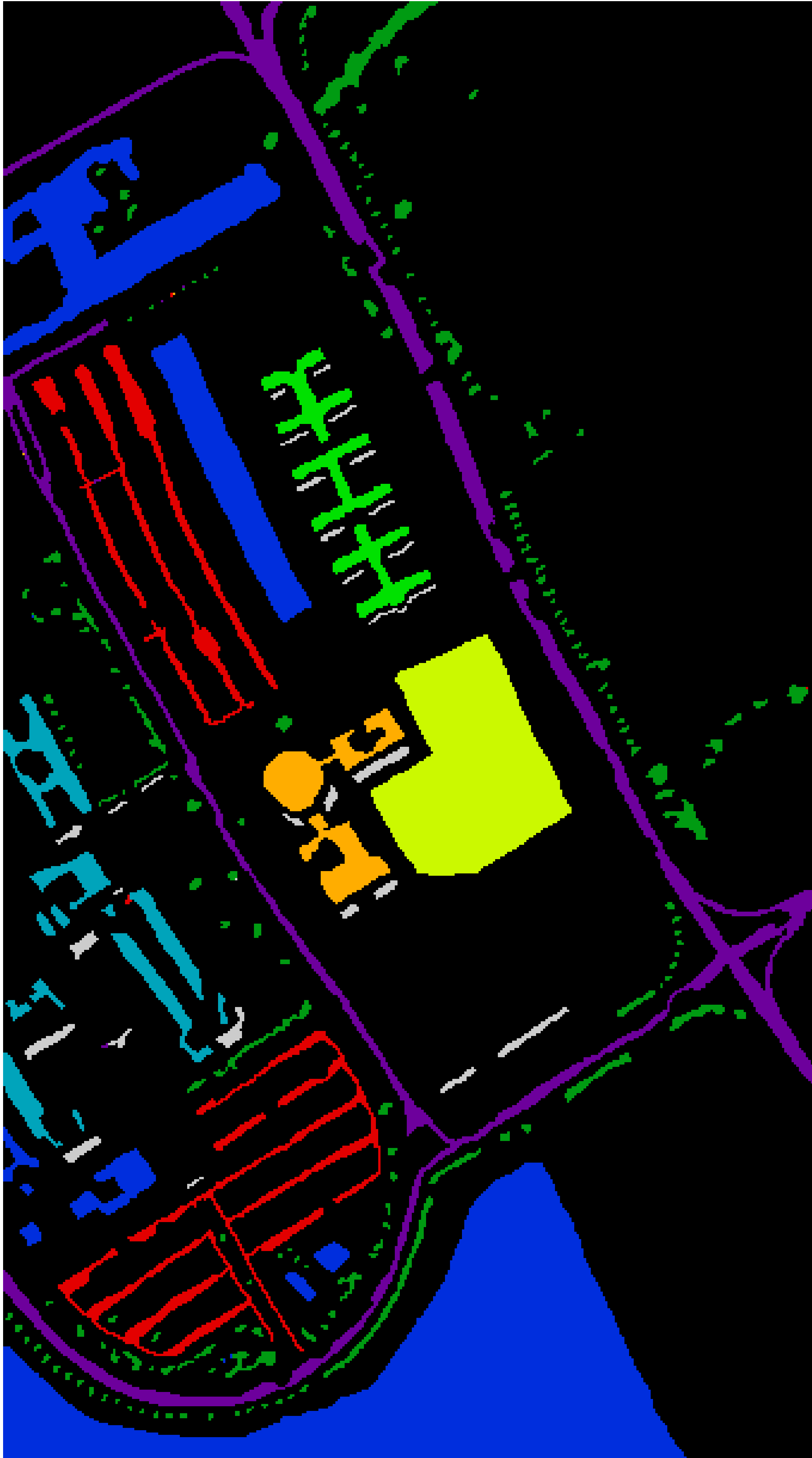}
		\caption{2D} 
		\label{Fig3A}
	\end{subfigure}
	\begin{subfigure}{0.070\textwidth}
		\includegraphics[width=0.99\textwidth]{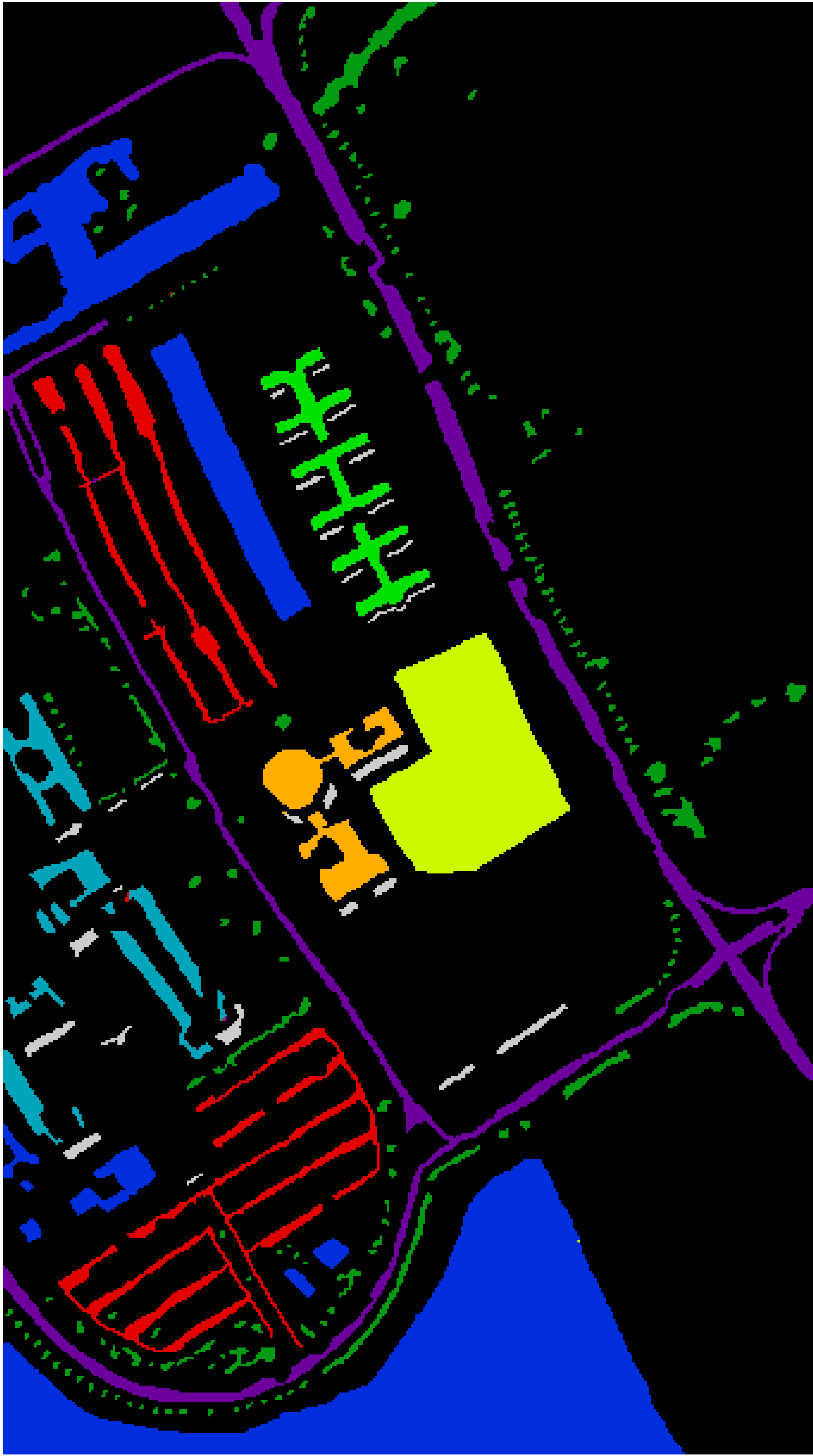}
		\caption{3D}
		\label{Fig3B}
	\end{subfigure}
		\begin{subfigure}{0.070\textwidth}
		\includegraphics[width=0.99\textwidth]{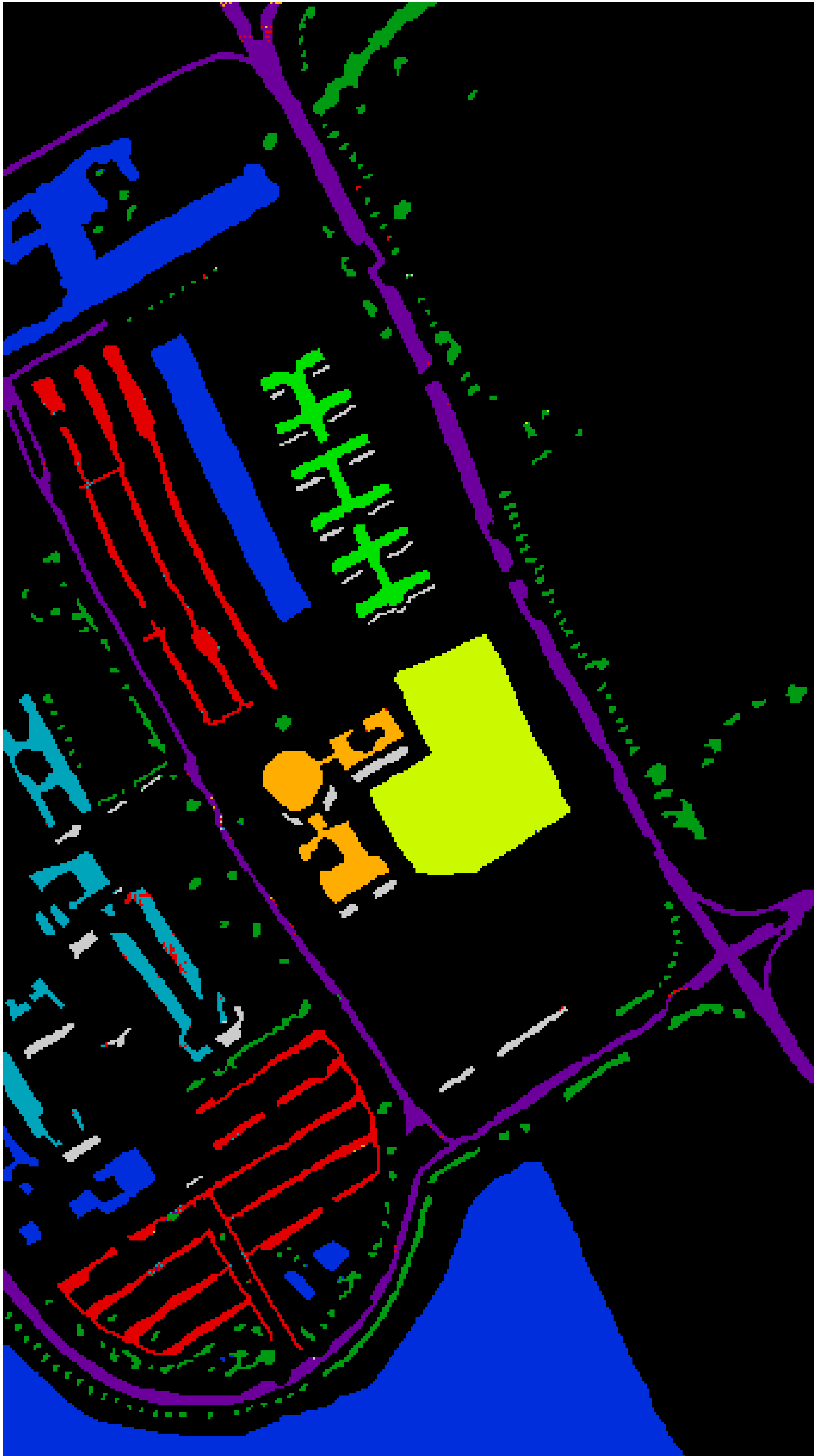}
		\caption{A.Net} 
		\label{Fig3C}
	\end{subfigure}
	\begin{subfigure}{0.070\textwidth}
		\includegraphics[width=0.99\textwidth]{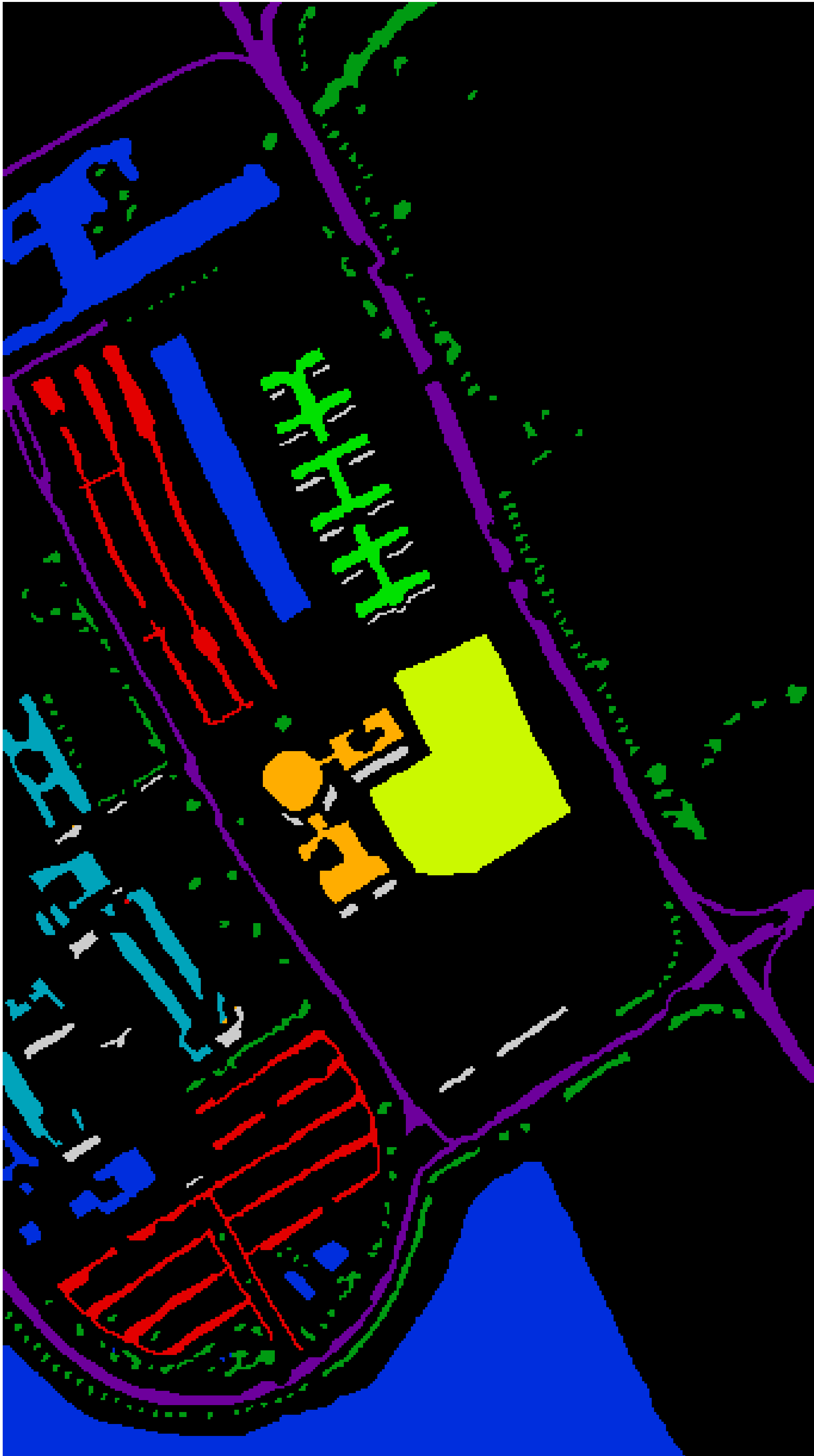}
		\caption{LeNet}
		\label{Fig3D}
	\end{subfigure}
		\begin{subfigure}{0.070\textwidth}
		\includegraphics[width=0.99\textwidth]{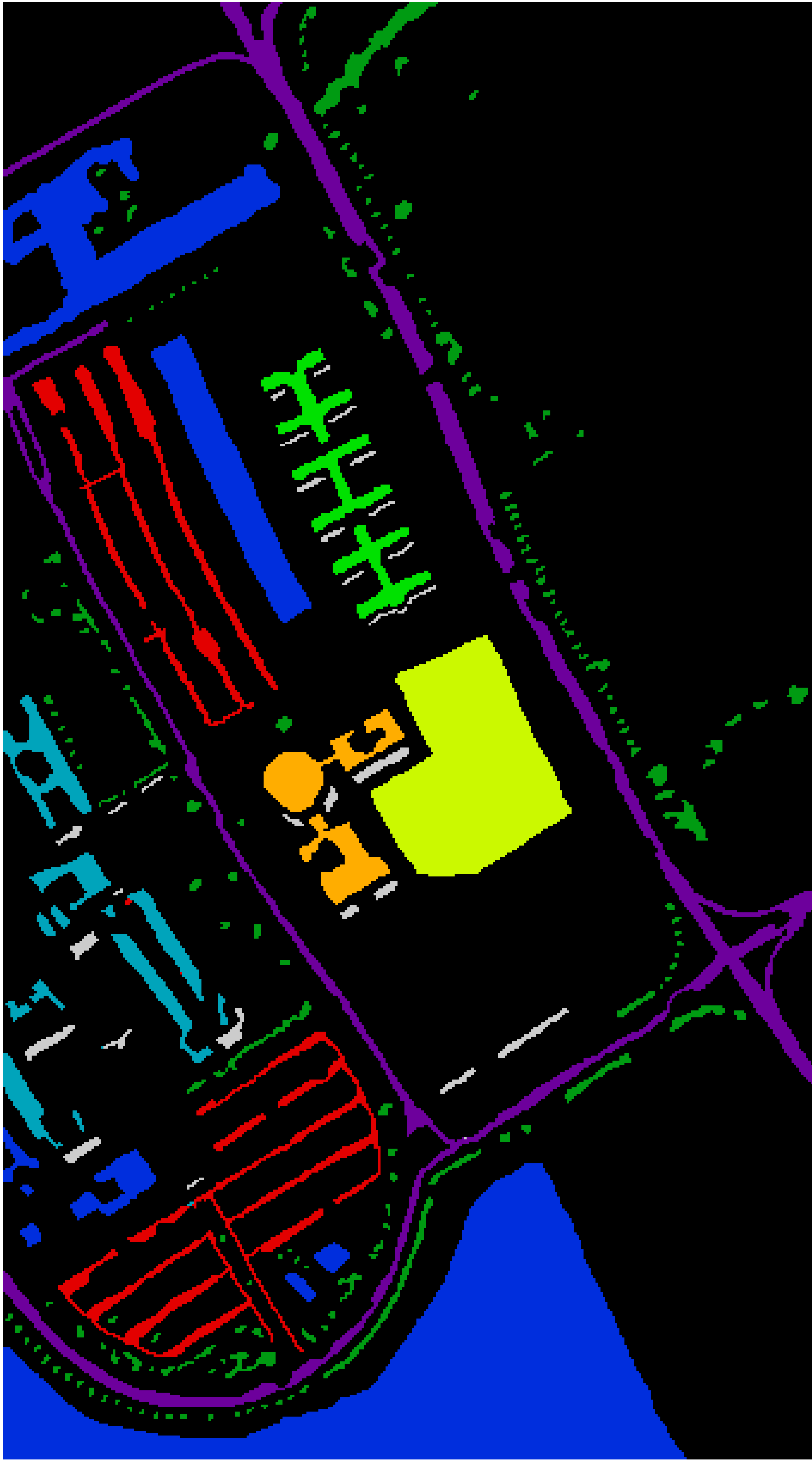}
		\caption{WoS} 
		\label{Fig3E}
	\end{subfigure}
	\begin{subfigure}{0.070\textwidth}
		\includegraphics[width=0.99\textwidth]{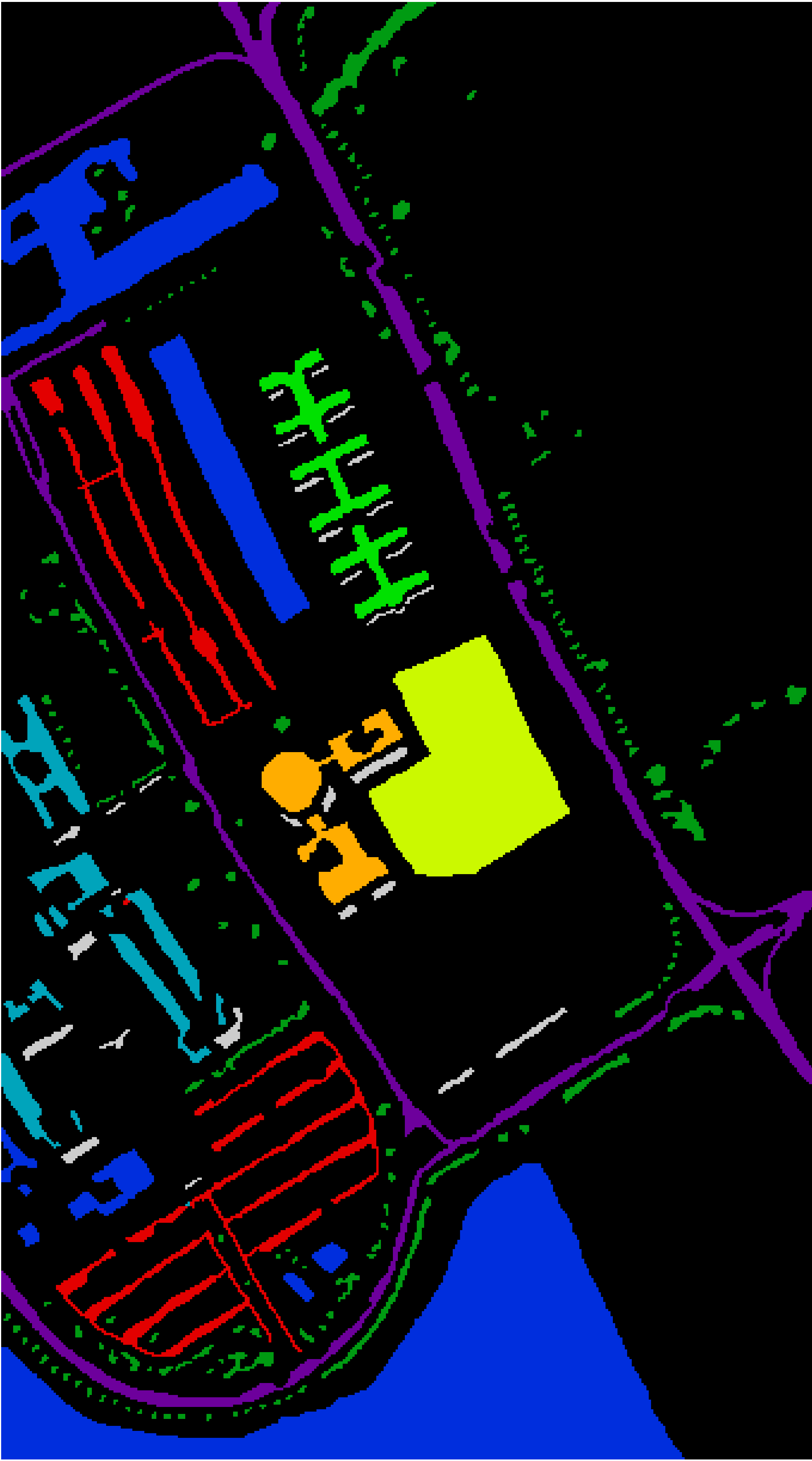}
		\caption{WS}
		\label{Fig3F}
	\end{subfigure}
\caption{\textbf{Pavia University:} Classification accuracy: Fig. \ref{Fig3A}: 2D-CNN = 99.9070\%, Fig. \ref{Fig3B}: 3D CNN = 99.9256\%, Fig. \ref{Fig3C}: AlexNet = 99.0768\%, Fig. \ref{Fig3D}: LeNet = 99.9318\%, Fig. \ref{Fig3E}: WoS (Hybrid net without Smoothing) = 99.9504\%, and Fig. \ref{Fig3F}: WS (with Smoothing) = \textbf{99.9628\%}.}
\label{Fig.3}
\end{figure}

Pavia University (PU) dataset acquired using Reflective Optics System Imaging Spectrometer (ROSIS) optical sensor over Pavia in northern Italy. PU dataset is distinguished into 9 different classes. PU consists of $610 \times 610$ spatial samples per spectral band and 103 spectral bands with a spatial resolution of 1.3m. Further details about the experimental datasets can be found at \cite{Grupointel}. Table \ref{Tab.2} and Figure \ref{Fig.3} presents in-depth comparative accuracy analysis on the IP dataset.

\begin{table}[!hbt]
    \centering
    \caption{\textbf{Pavia University:} Performance analysis of different state-of-the-art models. The higher accuracies are emphasised.}
    \resizebox{\columnwidth}{!}{
    \begin{tabular}{cccccccc} \\ \hline 
        \textbf{Class} & \textbf{Train/Val/Tesst} & \textbf{2D} & \textbf{3D} & \textbf{AlexNet} & \textbf{LeNet} & \textbf{WoS} & \textbf{WS} \\ \hline 
        Asphalt & 1658/1658/3316 & 100 & 100 & 98.9143 & 100 & 100 & \textbf{100}\\
        Meadows & 4662/4662/9324 & 100 & 99.9892 & 100 & 100 & 100 & \textbf{100}\\
        Gravel & 524/525/1049 & 99.5233 & 99.3326 & 95.5195 & 99.4280 & 99.5233 & \textbf{99.6186}\\
        Trees & 766/766/1532 & 99.6736 & 100 & 98.8250 & 99.7389 & 100 & \textbf{100}\\
        Painted & 336/337/673 & 100 & 100 & 100 & 100 & 100 & \textbf{100}\\
        Soil & 1257/1257/2514 & 100 & 100 & 99.9602 & 100 & 100 & \textbf{100}\\
        Bitumen & 332/333/665 & 100 & 100 & 99.6992 & 100 & 100 & \textbf{100}\\
        Bricks & 920/921/1841 & 99.8913 & 99.8913 & 97.9359 & \textbf{100} & 99.8913 & 99.8913\\
        Shadows & 237/237/74 & 99.3670 & 99.5780 & 98.5232 & 99.7890 & 99.7890 & \textbf{100}\\ \hline
        \multicolumn{2}{c}{\textbf{Training Time}} & 296.0174 & 1145.9233 & 4716.4900 & 308.6389 & 1143.5753 & 1143.4996\\ \hline 
        \multicolumn{2}{c}{\textbf{Test Time}} & 5.7400 & 14.8634 & 24.1701 & 4.5054 & 15.0601 & 15.5904 \\ \hline 
        \multicolumn{2}{c}{\textbf{Overall Accuracy}} & 99.9298 & 99.9438 & 99.3033 & 99.9485 & 99.9625 & \textbf{99.9719}\\ \hline
        \multicolumn{2}{c}{\textbf{Average Accuracy}} & 99.8283 & 99.8657 & 98.8197 & 99.8839 & 99.9115 & \textbf{99.9455}\\ \hline
        \multicolumn{2}{c}{\textbf{Kappa ($\kappa$)}} & 99.9070 & 99.9256 & 99.0768 & 99.9318 & 99.9504 & \textbf{99.9628}\\ \hline
    \end{tabular}}
    \label{Tab.2}
\end{table}

\section{Conclusion}
\label{Sec.4}

The paper proposed a 3D/2D regularized CNN feature hierarchy for HSIC, in which the loss contribution is considered as a mixture of entropy between a predicted distribution and the hot-encoding, and the entropy between the predicted and noise distribution. Several other regularization techniques (e.g., dropout, L1, L2, etc.) have also been used, however, these techniques to some extent, lead to predicting the samples extremely confidently which is not good from a generalization point of view. Therefore, this work proposed the use of an entropy-based regularization process to improve the generalization performance using soft labels. These soft labels are the weighted average of the hard labels and uniform distribution over entire ground truths. The entropy-based regularization process prevents CNN from becoming over-confident while learning and predicting thus improves the model calibration and bean-search. Extensive experiments have confirmed that the proposed pipeline outperformed several state-of-the-art methods.

{\footnotesize
\bibliographystyle{IEEEtran}
\bibliography{Sample}}
\end{document}